\documentclass[10pt,twocolumn,letterpaper]{article}

\usepackage{cvpr}              

\usepackage{multirow}

\usepackage[dvipsnames]{xcolor}

\usepackage{amsmath,amsfonts,bm}

\newcommand{\diff}{\mathop{}\!\mathrm{d}}

\def\eqref#1{equation~\ref{#1}}

\def\1{\bm{1}}

\DeclareMathAlphabet{\mathsfit}{\encodingdefault}{\sfdefault}{m}{sl}
\SetMathAlphabet{\mathsfit}{bold}{\encodingdefault}{\sfdefault}{bx}{n}

\newcommand{\R}{\mathbb{R}}

\definecolor{cvprblue}{rgb}{0.21,0.49,0.74}
\usepackage[pagebackref,breaklinks,colorlinks,citecolor=cvprblue]{hyperref}
\usepackage[accsupp]{axessibility} 

\def\paperID{113} 
\def\confName{CVPR}
\def\confYear{2024}

\title{Equipping Diffusion Models with Differentiable Spatial Entropy for\\Low-Light Image Enhancement}

\author{Wenyi Lian\\
Uppsala University\\
Sweden\\
{\tt\small shermanlian@163.com}
\and
Wenjing Lian\\
Northeastern University\\
China\\
{\tt\small abbylian04@gmail.com}
\and
Ziwei Luo\\
Uppsala University\\
Sweden\\
{\tt\small ziwei.luo@it.uu.se}
}

\begin{document}
\maketitle
\begin{abstract}
Image restoration, which aims to recover high-quality images from their corrupted counterparts, often faces the challenge of being an ill-posed problem that allows multiple solutions for a single input. However, most deep learning based works simply employ $\ell_1$ loss to train their network in a deterministic way, resulting in over-smoothed predictions with inferior perceptual quality. In this work, we propose a novel method that shifts the focus from a deterministic pixel-by-pixel comparison to a statistical perspective, emphasizing the learning of distributions rather than individual pixel values. The core idea is to introduce spatial entropy into the loss function to measure the distribution difference between predictions and targets. To make this spatial entropy differentiable, we employ kernel density estimation (KDE) to approximate the probabilities for specific intensity values of each pixel with their neighbor areas. Specifically, we equip the entropy with diffusion models and aim for superior accuracy and enhanced perceptual quality over $\ell_1$ based noise matching loss. In the experiments, we evaluate the proposed method for low light enhancement on two datasets and the NTIRE challenge 2024. All these results illustrate the effectiveness of our statistic-based entropy loss. Code is available at \url{https://github.com/shermanlian/spatial-entropy-loss}.
\end{abstract}    
\section{Introduction}
\label{sec:intro}

The quest for high-quality image restoration has led to advancements in neural networks and objective functions. Even so, achieving perceptually convincing restorations remains a challenge. Traditional pixel-wise loss functions, such as $\ell_1$ and $\ell_2$, fall short in capturing the perceptual qualities of images, often resulting in high fidelity (reflected in high PSNR scores) but over-smoothed outputs~\citep{ledig2017photo,lugmayr2020srflow,zhang2018unreasonable}. This highlights the necessity for methods that extend beyond pixel accuracy to embrace a broader spectrum of image attributes.

\begin{figure}[t]
  \centering
   \includegraphics[width=.9\linewidth]{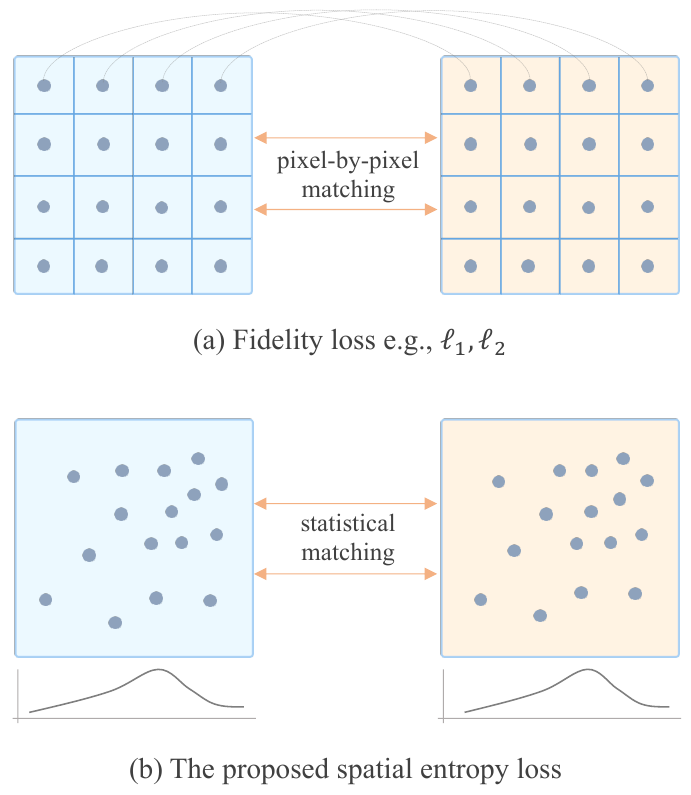}
   \caption{Illustration of the traditional fidelity loss (a) and the proposed spatial entropy loss (b). The fidelity loss uses pixel-by-pixel matching for two images while the proposed entropy loss adapts a statistical matching strategy for realistic image generation.}
   \label{fig:teaser}
\end{figure}

To address this issue, some studies utilize generative adversarial networks (GANs)~\citep{goodfellow2020generative} and perceptual metrics (e.g., VGG loss~\citep{johnson2016perceptual} and LPIPS loss~\citep{zhang2018unreasonable}) to improve the visual quality. While these methods have shown promise in generating more realistic images, they still rely on the $\ell_1$ loss to maintain the restoration accuracy. Moreover, both GAN and these perceptual metrics need additional (pre-trained) networks and are sensitive to different datasets and tasks. This not only increases the complexity of the model but also introduces a level of unpredictability, pointing to a need for more adaptable and self-sufficient solutions. 

Given the limitations of GANs and perceptual metrics, the traditional SSIM~\citep{wang2004image} offers a promising alternative by evaluating image similarity through statistical attributes like means and variances of local patches, rather than relying on direct pixel-wise distances. While SSIM is often considered superior to PSNR ($\ell_2$-based metric) in image quality assessment, its reliance on basic statistical measures (mean and variance only) limits its effectiveness. When used as a loss function, SSIM may also produce over-smoothed images, akin to the limitations observed with $\ell_1$ or $\ell_2$ loss due to its restricted spatial information.

In this work, our objective is to design a novel statistic-based loss function which measures the distribution similarity between predicted images and ground truth images, as shown in \Cref{fig:teaser}. The simplest way to represent an image distribution is through a 1D histogram, which calculates the probability of all pixel intensity values. This representation can then be utilized to compute meaningful statistic metrics such as the Shanno entropy~\citep{shannon1948mathematical} or relative entropy (KL-divergence) between two images. However, the limitation of a 1D histogram lies in its inability to comprehensively capture spatial information, thus rendering it inadequate for representing the entirety of an image. Additionally, the fact that the counting operation in histogram calculation is not differentiable prevents its use in training within a deep learning framework.

To address the above problems, this work proposes a differentiable spatial entropy. This method utilizes neighboring pixels to gather extra spatial information and employs kernel density estimation (KDE) to ensure the differentiability of the histogram/entropy. Moreover, to enhance the richness of pixel neighbor information, we suggest incorporating of randomly shuffled weights, which also improves the robustness of distribution matching. In experiments, we apply our entropy loss to diffusion models for low light image enhancement. The results shows that our new loss can preserve image restoration performance (e.g., PSNR/SSIM metrics) and further enhance the visual quality of predictions  (in LPIPS~\citep{zhang2018unreasonable}, FID~\citep{heusel2017gans} scores). 

Our main contributions are summarized as follows:

\begin{enumerate}
    \item We present a novel statistic-based objective that optimizes distribution similarity rather than pixel-wise distance. Compared to other commonly used loss functions such as $\ell_1$, $\ell_2$, or GAN loss, it provides a new perspective for generative modelling.
    \item We leverage KDE to make this spatial entropy differentiable. By simply equipping it with diffusion models for noise matching, the diffusion performance for realistic image generation is significantly improved.
    \item Extensive experiments on the low light enhancement task demonstrate that the proposed entropy is effective for diffusion-based image generation.
    
\end{enumerate}





\section{Preliminaries}
\subsection{Image 1D Entropy} Consider an image $X$ of size $H \times W$, with pixel intensities as non-negative integers from $0$ to $L$. The 1D Shannon entropy~\citep{shannon1948mathematical} $H(\cdot)$ is defined as:
\begin{equation}
    \centering
    \displaystyle H(X) = -\sum_{i=0}^{L} p_i \log(p_i),
    \label{eq:en1d}
\end{equation}
where $L$ is the maximum grey level and $p_i$ is the probability of $i$-th intensity value given by: 
\begin{equation}
    \centering
    \displaystyle p_i = \frac{\sum_{x \in X} \mathbb{I}\{x=i\}}{H \times W},
    \label{eq:pi}
\end{equation}

where $\mathbb{I}\{x=i\}$ is the indicator function that counts the occurrence of pixels with intensity $i$, and the denominator $H \times W$ represents the total number of pixels in the image, with $H$ being the height and $W$ the width of the image. 

\cref{eq:pi} calculates the frequency of each intensity level $i$ through a counting operation, effectively capturing the intensity distribution across an image. However, it overlooks the spatial arrangement and the relationships between pixels, which can be valuable for uncovering the deeper context and structural details of the image.

\begin{figure*}[t]
  \centering
   \includegraphics[width=.7\linewidth]{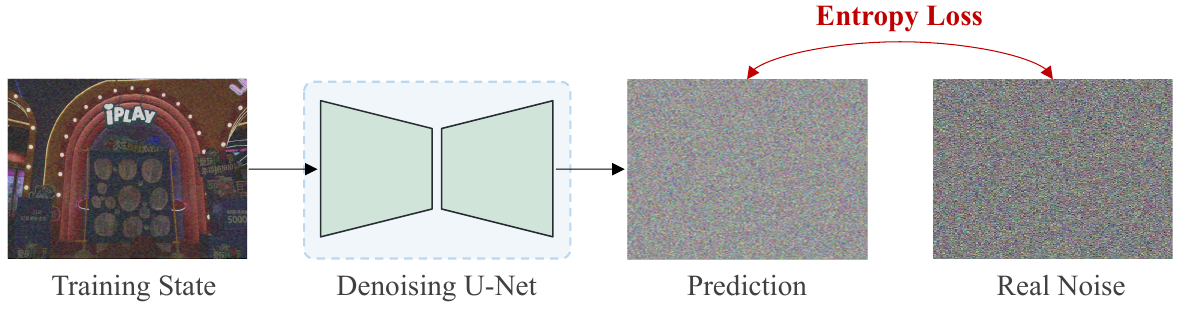}
   \caption{Overview of the training process for a diffusion model with the proposed spatial entropy loss for noise matching.}
   \label{fig:training_overview}
\end{figure*}

\subsection{Image Spatial Entropy} To enhance the representation of an image, spatial entropy considers additional spatial characteristics of pixels by examining their immediate surroundings~\citep{901061,413562}. Specifically, for any given pixel \(x\), we define \(\mathbb{N}\) as the set of its 8 neighboring pixel values. And $\tilde{x} \in \Omega$ is the rounded value of the mean of $\mathbb{N}$. Spatial entropy is then calculated as follows:
\begin{equation}
    \centering
    \displaystyle H_s(X) = -\sum_{i=0}^L \sum_{j \in \Omega} p_{i,j}\log p_{i,j},
    \label{eq:en2d}
\end{equation}
where $p_{i,j}$ denotes the probability of occurrence for a tuple of values $(i,j)$, where $i$ represents the intensity value of a current pixel and $j$ represents the average intensity value of its surrounding eight neighbors \(\mathbb{N}\). Similar to \cref{eq:pi}, $p_{i,j}$ is obtained by: 
\begin{equation}
    \centering
    \displaystyle p_{i, j} = \frac{\sum_{x \in X} \mathbb{I}\{x=i\ \, \& \, \tilde{x}=j\}}{H \times W}.
    \label{eq:pij}
\end{equation}

Compared to 1D entropy, which only considers the frequency of individual pixel intensities, the calculation of spatial entropy (\cref{eq:pij}) incorporates spatial relationships by statistically evaluating the mean intensity of pixels in a neighborhood. This approach enables spatial entropy to more accurately represent an image's spatial distribution, offering a deeper insight into its textural and structural complexity.

However, it is worth noting that both \cref{eq:pi} and \cref{eq:pij} contain a non-differentiable indicator function, which means we cannot use them as intermediate steps in the training of neural networks.

\subsection{Diffusion Model}
\label{sec:pre_diffusion}

The general diffusion model consists of a forward process and a backward process. Specifically, given the input variable $x_0$, the forward process iteratively adds noise to it with satisfying
\begin{equation}
    q(x_t \mid x_{t-1}) = \mathcal{N}(x_t; \sqrt{1 - \beta_t} x_{t-1}, \beta_t I),
\end{equation}
where $\beta_t$ is the predefined variance. With the nice property of Gaussian, we could have
\begin{equation}
    q(x_t \mid x_0) = \mathcal{N}(x_t; \sqrt{\bar{\alpha}_t} x_0, (1 - \bar{\alpha}_t) I),
\end{equation}
where $\alpha_t = 1 - \beta_t$ and $\bar{\alpha}_t = \prod_{s=1}^t \alpha_s$. Note that as $t \to \infty$, the sample of $q(x_t \mid x_0)$ converges to a Gaussian noise $\mathcal{N}(0, I)$. Then the backward process reverses the noise to the initial variable $x_0$, which naturally models the generative process and can reconstruct high-quality images.

Importantly, the diffusion forward process can be also modeled as a forward Stochastic Differential Equation (SDE) with continuous variables, defined by
\begin{equation}
	\diff {x} = f({x}, t) \diff t + g(t)\diff w, \quad {x}(0) \sim p_0({x}), 
	\label{eq:sde}
\end{equation}
where $f$ and $g$ are the drift and dispersion functions, respectively, $w$ is a standard Wiener process, and ${x}(0) \in \R^{d}$ is an initial condition. And similarly, such an SDE can be reversed to model the generative process:
\begin{equation}
        \diff {x} = \Bigl[ f({x}, t) - g(t)^2\, \nabla_{{x}} \log p_t({x}) \Bigr] \diff t + g(t) \diff \hat{w}, 
    \label{eq:reverse-sde}
\end{equation}
where ${x}(T) \sim p_T({x})$ and $\hat{w}$ is a reverse-time Wiener process and $p_t({x})$ stands for the marginal probability density function of ${x}(t)$ at time~$t$. The score function $\nabla_{x} \log p_t({x})$ is in general intractable and thus SDE-based diffusion models approximate it by training a time-dependent neural network $s_\theta({x}, t)$ under a so-called score matching objective.

\section{Method}

\subsection{Differentiable Spatial Entropy}

The idea is to introduce the kernel density estimation (KDE)~\cite{parzen1962estimation} to estimate the smooth probability density function (PDF) instead of directly counting the number of the same intensity value (\cref{eq:pi}) or tuple (\cref{eq:pij}).
Formally, we define the one-dimensional KDE for the probability of a specific intensity value $i$ as:
\begin{align}
    \centering
    \displaystyle p_i \approx \hat{f}_h(i) &= \frac{1}{2Nh} \sum_{x \in X} \mathbb{I}\left \{\frac{|x-i|}{h} < 1 \right \}\\
    &= \frac{1}{Nh} \sum_{x \in X} \mathcal{K}(\frac{x-i}{h}),
    \label{eq:kde1d}
\end{align}
where $N = H \times W$ is the total number of pixels, $\mathcal{K}(\cdot)$ represents the generalized kernel substituting the indicator function $\frac{1}{2}\mathbb{I}\{\cdot\}$ and $h$ is a smoothing parameter called the bandwidth. The above 1D KDE is derived from \cref{eq:pi} and it can be made differentiable by choosing an appropriate kernel function such as Gaussian distribution. 
Similarly, to estimate the spatial probability density for the tuple $(i, j)$, we rewrite \cref{eq:pij} to a spatial KDE for $p_{i, j}$ as follows:
\begin{align}
    \centering
    \displaystyle \hat{f}_h(i, j) &= \frac{1}{2Nh} \sum_{x \in X} \mathbb{I}\left \{\frac{|x-i|}{h}< 1 \And \frac{|\tilde{x}-j|}{h} < 1  \right \}  \\
    &=\frac{1}{2Nh} \sum_{x \in X} \mathbb{I}\left \{\frac{|x-i|}{h} < 1 \right \} \cdot \mathbb{I}\left \{ \frac{|\tilde{x}-j|}{h}< 1  \right \}.
    \label{eq:kde2d}
\end{align}
By replacing the indicator function $\frac{1}{2}\mathbb{I}\{\cdot \}$ with $\mathcal{K}(\cdot)$, we have the following:
\begin{align}
    \centering
    \displaystyle \hat{f}_h(i, j) 
        &=\frac{2}{Nh} \sum_{x \in X} \frac{1}{2} \mathbb{I}\left \{ \frac{|x-i|}{h} < 1 \right \} \cdot \frac{1}{2}\mathbb{I}\left \{\frac{|\tilde{x}-j|}{h} < 1  \right \} \\
        &=\frac{2}{Nh } \sum_{x \in X} \mathcal{K}_1(\frac{x-i}{h}) \cdot \mathcal{K}_2(\frac{\tilde{x}-j}{h} )
    \label{eq:kde2d_K}
\end{align}
Here $\mathcal{K}_1$ and $\mathcal{K}_2$ could be the same kernel function. 
In particular, valid kernels should be real-valued, non-negative, symmetric, and satisfy the normalization: $\int \mathcal{K}(t)d{t} = 1$, which guarantees the spatial KDE is a probability density function. We usually choose Gaussian or the derivative of sigmoid~\cite{avi2023differentiable} as the kernel function.

\subsection{Spatial Entropy Loss for Image Restoration}

To make use of the differentiable spatial KDE in image generation, we introduce statistical losses to measure the distribution distance between the predicted image and ground truth. A typical example is the relative entropy (also called KL-divergence). Mathematically, let $P$ and $Q$ represent the ground truth's and prediction's probability distributions computed from \cref{eq:kde2d_K}, the spatial relative entropy is defined as follows:
\begin{equation}
    \centering
    \displaystyle H_s(P, Q) = D_{KL}(P||Q) = -\sum_{i=0}^L \sum_{j \in \Omega} p_{i,j}\log \frac{q_{i,j}}{p_{i,j}}.
    \label{eq:skl}
\end{equation}
Note that the choice of the loss function is flexible, other statistic measurements like Cross-entropy and Hellinger distance~\cite{van2000asymptotic} can also be used in our framework. In addition, although we directly employ spatial relative entropy as our loss function for network training, the statistics of the whole image are too global and spatially non-stationary, which cannot guide the reconstruction of local details. To address this issue, we further propose to compute the relative entropy locally to provide a spatially varying quality map of the image, which is more stable and efficient for different image distortions. Similar to SSIM~\cite{wang2004image}, we slide a $11 \times 11$ square window to compute the local probabilities, which can be regarded as a convolution operation by setting all weights to one and then replace the $N$ with the total number of pixels i.e. $11 \times 11$ in \cref{eq:kde2d_K}.

\paragraph{Shuffle Weights for Neighborhoods Augmentation}
In previous works, the mean value of neighborhood pixels often serves as the spatial information. However, only using one pattern of averaging neighborhoods is still not sufficient to represent the whole image. To better improve the spatial information during training, we propose adding random weights to neighboring pixels for each entropy calculation, as a strategy for weighted averaging augmentation. In theory, this approach of differentiable spatial entropy, enhanced by the random weights strategy, can be more broadly and effectively applied across various image generation frameworks and applications.

\paragraph{Equipping Entropy to Diffusion Models}
Most diffusion models adopt noise-matching loss (NML) for generative modelling. In particular, NML uses $\ell_2$ to match predicted noise and a real Gaussian noise at a random sampled time $t$. To equip the proposed entropy to common diffusion models, we just simply replace the $\ell_2$ term in noise matching. In this paper, we use Refusion~\cite{luo2023refusion} as the base diffusion framework which learns a mean-reverting SDE~\cite{luo2023image} (a special case in \cref{eq:sde}) for image restoration. \Cref{fig:training_overview} illustrated the combination of diffusion models with entropy loss. To further improve the distortion performance (in terms of PSNR and SSIM metrics), we follow Refusion to change NML to the maximum likelihood loss (MLL)~\cite{luo2023image} for optimal path reversing.

\section{Experiments}
In this section, we provide evaluations on the low light enhancement task with two datasets from LoL-v1 and LoL-v2 in \cref{sec:comparison} and analyze the effectiveness of the entropy in \cref{sec:ablation}. Moreover, in \cref{sec:ntire}, we further report the final results of the corresponding NTIRE 2024 low light enhancement challenge. Finally, the limitations and future work are presented in \cref{sec:limitation}.

\subsection{Datasets and Implementation Details}
\label{sec:implementation}

\paragraph{Datasets.} Our method is evaluated on the Low Light Paired (LoL) datasets: LOL-v1~\cite{wei2018deep} and LOL-v2-real~\cite{yang2020fidelity}.
The LOL-v1 dataset is the first of its kind, comprising image pairs captured from real-world scenarios. It consists of 500 pairs of low-light and normal-light images, with 485 pairs for training and 15 for testing. These images maintain a consistent resolution of 400×600 pixels and mostly depict indoor scenes with typical low-light noise.
In contrast, the LOL-v2-real dataset contains 689 pairs of low-/normal-light images for training and additional 100 pairs for testing. This dataset captures a diverse range of scenes, both indoors and outdoors, under varying lighting conditions. The diversity of the LOL-v2-real dataset allows us to learn from a broader spectrum of low-light scenarios, enhancing its evaluation of image enhancement techniques.

\paragraph{Implementation Details.} We develop the image restoration SDE (IR-SDE)~\cite{luo2023image} model targeting low-light image enhancement with a Conditional NAFNet~\cite{chen2022simple} architecture. This model employs the AdamW optimizer with \(\beta_1 = 0.9\) and \(\beta_2 = 0.99\) over \(200,000\) iterations. The initial learning rate is \(5 \times 10^{-5}\), reduced to \(1 \times 10^{-6}\) through a cosine annealing schedule. Training samples are processed in batches of 8, with 128x128 patches randomly cropped from low-/normal-light image pairs. Data augmentation includes random rotation and flipping. In addition, we set the SDE timestep to 100 in both training and testing. All our experiments are implemented using the PyTorch framework and performed on an NVIDIA A100 GPU with 2 days of training for each dataset.

\begin{table}[t]
\caption{Quantitative comparison between the proposed method with other state-of-the-art low-light enhancement approaches on the LOL-v1~\cite{wei2018deep} test set. The best results are marked in \textbf{bold}.}
\label{table:lolv1}
\begin{center}
\resizebox{1.\linewidth}{!}{
\begin{tabular}{lcccc}
\toprule
\multirow{2}{*}{Method} &  \multicolumn{2}{c}{Distortion} & \multicolumn{2}{c}{Perceptual}  \\ \cmidrule(lr){2-3} \cmidrule(lr){4-5}
&  PSNR$\uparrow$ & SSIM$\uparrow$ & LPIPS$\downarrow$ & FID$\downarrow$   \\
\midrule

NPE~\cite{wang2013naturalness}  & 16.97 & 0.484 & 0.400 & 104.1  \\
SRIE~\cite{fu2016weighted}  & 11.86 & 0.495 & 0.353 & 88.73  \\
LIME~\cite{guo2016lime}  & 17.55 & 0.531 & 0.387 & 117.9  \\
RetinexNet~\cite{wei2018deep}  & 16.77 & 0.462 & 0.417 & 126.3  \\
DSLR~\cite{lim2020dslr}  & 14.82 & 0.572 & 0.375 & 104.4  \\
DRBN~\cite{yang2020fidelity}  & 16.68 & 0.730 & 0.345 & 98.73  \\
Zero-DCE~\cite{guo2020zero}  & 14.86 & 0.562 & 0.372 & 87.24  \\
MIRNet~\cite{zamir2020learning}  & 24.14 & 0.830 & 0.250 & 69.18  \\
EnlightenGAN~\cite{jiang2021enlightengan}  & 17.61 & 0.653 & 0.372 & 94.70  \\
ReLLIE~\cite{zhang2021rellie}  & 11.44 & 0.482 & 0.375 & 95.51  \\
RUAS~\cite{liu2021retinex}  & 16.41 & 0.503 & 0.364 & 102.0  \\
DDIM~\cite{song2020denoising}  & 16.52 & 0.776 & 0.376 & 84.07  \\
CDEF~\cite{lei2022low}  & 16.34 & 0.585 & 0.407 & 90.62  \\
SCI~\cite{ma2022toward}  & 14.78 & 0.525 & 0.366 & 78.60  \\
URetinex-Net~\cite{wu2022uretinex}  & 19.84 & 0.824 & 0.237 & 52.38  \\
SNRNet~\cite{xu2022snr}  & 24.61 & 0.842 & 0.233 & 55.12  \\
Uformer~\cite{wang2022uformer}  & 19.00 & 0.741 & 0.354 & 109.4  \\
Restormer~\cite{zamir2022restormer}  & 20.61 & 0.797 & 0.288 & 73.00  \\
Palette*~\cite{saharia2022palette}  & 11.77 & 0.561 & 0.498 & 108.3  \\
$\mathrm{UHDFour}_{2\times}$~\cite{li2023embedding}  & 23.09 & 0.821 & 0.259 & 56.91  \\
WeatherDiff~\cite{ozdenizci2023restoring}  & 17.91 & 0.811 & 0.272 & 73.90  \\
GDP~\cite{fei2023generative}  & 15.90 & 0.542 & 0.421 & 117.5  \\
Retinexformer~\cite{Cai_2023_ICCV}  & 25.16 & 0.845 & 0.130 & 71.21  \\
DiffLL~\cite{jiang2023low}  & \textbf{26.34} & 0.845 & 0.217 & 48.11  \\
Entropy-SDE (\textbf{Ours})  & 24.05 & \textbf{0.848} & \textbf{0.081} & \textbf{37.20} \\

\bottomrule
\end{tabular}
}
\end{center}
\vskip -0.1in
\end{table}

\begin{table}[t]
\renewcommand{\arraystretch}{1.0}
\caption{Quantitative comparison between the proposed method with other state-of-the-art low-light enhancement approaches on the LOL-v2-real~\cite{yang2020fidelity} test set. The best results are marked in \textbf{bold}.}
\label{table:lolv2}
\begin{center}
\resizebox{1.\linewidth}{!}{
\begin{tabular}{lcccc}
\toprule
\multirow{2}{*}{Method} &  \multicolumn{2}{c}{Distortion} & \multicolumn{2}{c}{Perceptual}  \\ \cmidrule(lr){2-3} \cmidrule(lr){4-5}
&  PSNR$\uparrow$ & SSIM$\uparrow$ & LPIPS$\downarrow$ & FID$\downarrow$   \\
\midrule

NPE~\cite{wang2013naturalness}  & 17.33 & 0.464 & 0.396 & 100.0  \\
SRIE~\cite{fu2016weighted}  & 14.45& 0.524 & 0.332 & 78.83    \\
LIME~\cite{guo2016lime}  &17.48 & 0.505 & 0.428 & 118.2    \\
RetinexNet~\cite{wei2018deep}   &17.72 & 0.652 & 0.436 & 133.9  \\
DSLR~\cite{lim2020dslr}  &17.00 & 0.596 & 0.408 & 114.3     \\
DRBN~\cite{yang2020fidelity} & 18.47 & 0.768 & 0.352 & 89.09  \\
Zero-DCE~\cite{guo2020zero}  &18.06 & 0.580 & 0.352 & 80.45  \\
MIRNet~\cite{zamir2020learning}  &20.02 & 0.820 & 0.233 & 49.10  \\
EnlightenGAN~\cite{jiang2021enlightengan}  &18.68 & 0.678 & 0.364& 84.04  \\
ReLLIE~\cite{zhang2021rellie}  & 14.40 & 0.536 & 0.334 & 79.84\\
RUAS~\cite{liu2021retinex}  &15.35 & 0.495 & 0.395 & 94.16\\
DDIM~\cite{song2020denoising}  & 15.28 & 0.788 & 0.387 & 76.39 \\
CDEF~\cite{lei2022low}  &19.76 & 0.630 & 0.349 & 74.06\\
SCI~\cite{ma2022toward}  &17.30 & 0.540 & 0.345 & 67.62\\
URetinex-Net~\cite{wu2022uretinex}  &21.09 & 0.858 & 0.208 &  49.84 \\
SNRNet~\cite{xu2022snr}  &21.48 & 0.849 & 0.237 &54.53   \\
Uformer~\cite{wang2022uformer}  &18.44 & 0.759 & 0.347& 98.14\\
Restormer~\cite{zamir2022restormer}  &24.91 & 0.851 & 0.264 & 58.65  \\
Palette~\cite{saharia2022palette}  &14.70& 0.692 & 0.333& 83.94 \\
$\mathrm{UHDFour}_{2\times}$~\cite{li2023embedding}  & 21.79 & 0.854 & 0.292 & 60.84 \\
WeatherDiff*~\cite{ozdenizci2023restoring}  &20.00 & 0.829 & 0.253 &59.67 \\
GDP~\cite{fei2023generative}  &14.29 & 0.493 & 0.435 & 102.4\\
Retinexformer~\cite{Cai_2023_ICCV}  &22.80 & 0.840 & 0.169 & 62.46 \\
DiffLL~\cite{jiang2023low}  & \textbf{28.86} & \textbf{0.876} & 0.207 & \textbf{45.36}  \\
Entropy-SDE (\textbf{Ours})  & 21.31& 0.832 & \textbf{0.120} & 49.61\\

\bottomrule
\end{tabular}
}
\end{center}
\vskip -0.1in
\end{table}

\subsection{Comparison with State-of-the-Arts}
\label{sec:comparison}

\paragraph{Comparison Methods}
In this section, we compare the proposed method with the following state-of-the-art methods: NPE~\cite{wang2013naturalness}, SRIE~\cite{fu2016weighted}, LIME~\cite{guo2016lime}, RetinexNet~\cite{wei2018deep}, DSLR~\cite{lim2020dslr}, DRBN~\cite{yang2020fidelity}, Zero-DCE~\cite{guo2020zero}, MIRNet~\cite{zamir2020learning}, EnlightenGAN~\cite{jiang2021enlightengan}, ReLLIE~\cite{zhang2021rellie}, RUAS~\cite{liu2021retinex}, DDIM~\cite{song2020denoising}, CDEF~\cite{lei2022low}, SCI~\cite{ma2022toward}, URetinex-Net~\cite{wu2022uretinex}, SNRNet~\cite{xu2022snr}, Uformer~\cite{wang2022uformer}, Restormer~\cite{zamir2022restormer}, Palette~\cite{saharia2022palette}, $\mathrm{UHDFour}_{2\times}$~\cite{li2023embedding}, WeatherDiff*~\cite{ozdenizci2023restoring}, GDP~\cite{fei2023generative}, Retinexformer~\cite{Cai_2023_ICCV}, DiffLL~\cite{jiang2023low}. These methods can be roughly split into four categories: optimization-based approaches, learning-based approaches, transformer-based approaches, and diffusion-based approaches. Our method belongs to the last category. For most methods, we report their results from the DiffLL~\cite{jiang2023low} paper. And we compare the visual results with EnlightenGAN~\cite{jiang2021enlightengan} and URetinex-Net~\cite{wu2022uretinex} by re-testing their official pretrained models.

\paragraph{Evaluation Metrics}
As the purpose of this work is to improve the visual quality of diffusion-based results in low-light enhancement, we thus focus more on the perceptual metrics: Learned Perceptual Image Patch Similarity (LPIPS)~\citep{zhang2018unreasonable} and Fr\'{e}chet inception distance (FID)~\citep{heusel2017gans}. In addition, distortion metrics like PSNR and SSIM~\cite{wang2004image} are also reported to measure the image fidelity for reference.

\begin{figure*}[t]
  \centering
   \includegraphics[width=1.\linewidth]{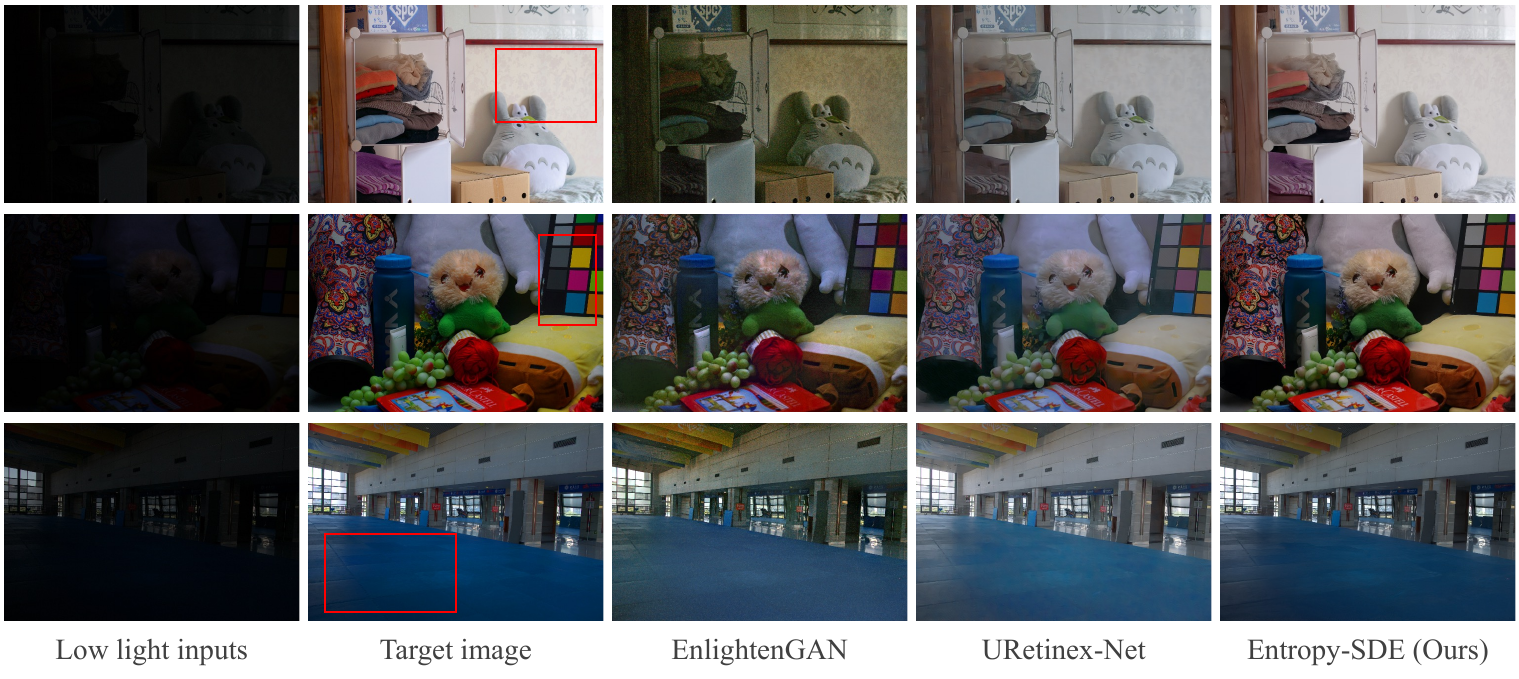}
   \caption{Visual comparison of the proposed model with other low light enhancement approaches on the LoL-v1~\cite{wei2018deep} dataset.}
   \label{fig:results_lolv1}
\end{figure*}

\begin{figure*}[t]
  \centering
   \includegraphics[width=1.\linewidth]{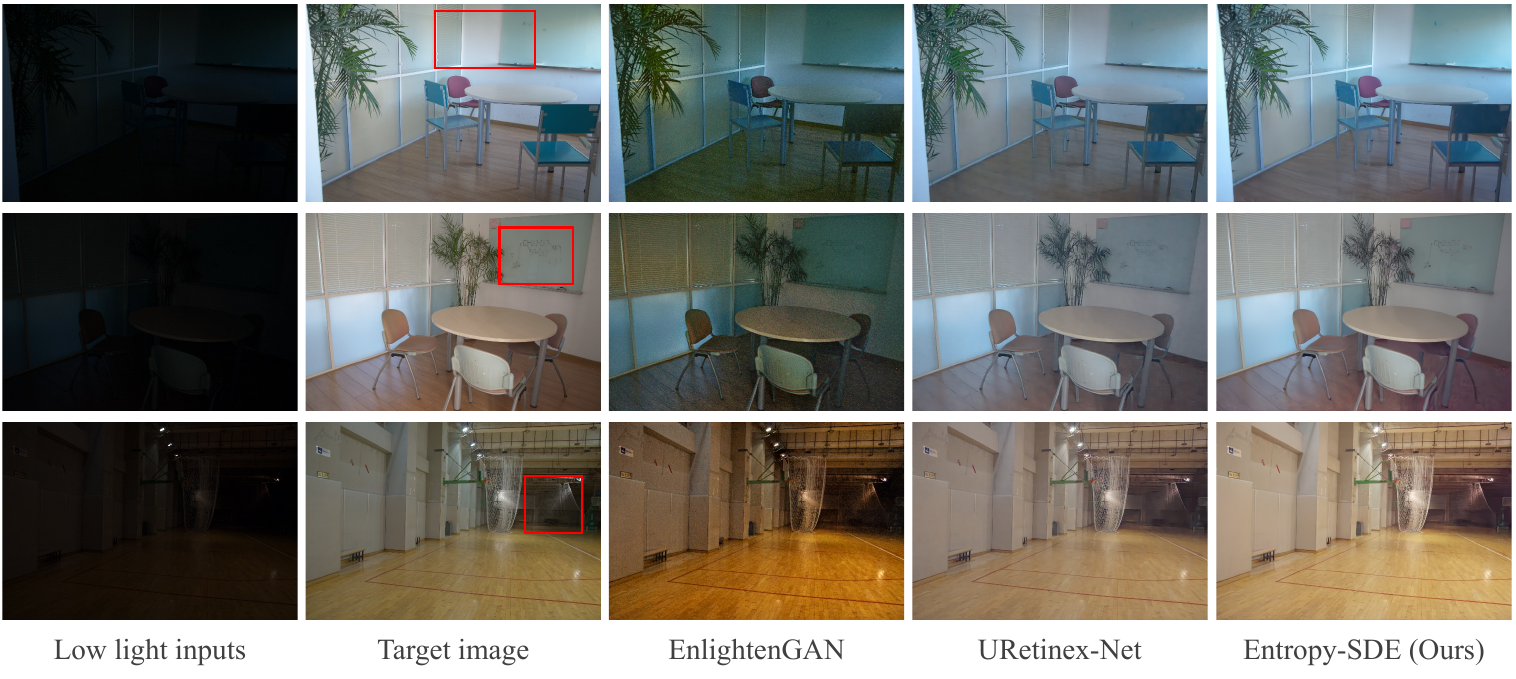}
   \caption{Visual comparison of the proposed model with other low light enhancement approaches on the LoL-v2-real~\cite{yang2020fidelity} dataset.}
   \label{fig:results_lolv2}
\end{figure*}

\paragraph{Results}
The quantitative comparison results on two datasets (LoL-v1~\cite{wei2018deep} and LoL-v2-real~\cite{yang2020fidelity}) are illustrated in \Cref{table:lolv1} and \Cref{table:lolv2}. It is observed that most advanced image restoration approaches (such as the MIRNet, SNRNet, Restormer, and Retinexformer) only perform well on the distortion performance (high PSNR and SSIM), meaning that they are easy to fit pixels but tend to produce relatively smooth or blurry results. DiffLL~\cite{jiang2023low} uses a specific wavelet-based diffusion model for low light enhancement and achieves the overall best results for both datasets. By leveraging the entropy loss in training, our method (Entropy-SDE) performs comparably with DiffLL on perceptual metrics in the LoL-v2-real dataset and even achieves the best performance over most metrics on the LoL-v1 dataset. These results illustrate that combining the proposed entropy loss with diffusion models works well on various datasets. In addition, it is reasonable that our fidelity results are inferior to other methods since we only leverage statistical information but PSNR is computed based on pixel-by-pixel distance.

The visual comparison of our method with EnlightenGAN~\cite{jiang2021enlightengan} and URetinex-Net~\cite{wu2022uretinex} on two datasets are shown in \Cref{fig:results_lolv1} and \Cref{fig:results_lolv2}, respectively. Note that EnlightenGAN always encounters the color shift problem due to the unstable GAN training process. The results produced by URetinex-Net look better in colors but lose some details in reconstruction (such as the blue point in the first-row of \Cref{fig:results_lolv2}). Our results are consistent with the target images while also look more realistic.

\subsection{Effectiveness of the Entropy Loss}
\label{sec:ablation}

In this section, we include ablation experiments further to analyze the effectiveness of the proposed spatial entropy Loss. In particular, the Refusion model is used as the baseline and we add individual $\ell_1$ loss and entropy to it. Here, we use a comprehensive set of metrics for evaluation, including PSNR, SSIM~\cite{wang2004image}, LPIPS~\citep{zhang2018unreasonable}, DISTS~\cite{ding2020image}, FID~\citep{heusel2017gans}, and NIQE~\cite{mittal2012making}. Note the NIQE is a non-reference metric that only measures the visual quality of outputs. In contrast, all other metrics measure both visual quality and the consistency between inputs and outputs.

The results on two datasets are shown in \Cref{table:entropy-lolv1} and \Cref{table:entropy-lolv2}. Obviously, baseline models with the proposed entropy loss can improve the results significantly. It is worth noting that the entropy loss only employs statistical information from images but still outperforms the $\ell_1$ loss version, further demonstrating the effectiveness of the proposed spatial entropy loss and the importance of differentiable solutions to the entropy.

\begin{table}[t]
\caption{Ablation experiment for the proposed entropy loss on LOL-v1~\cite{wei2018deep} test set. The Refusion~\cite{luo2023refusion} is used as the baseline.}
\label{table:entropy-lolv1}
\begin{center}
\resizebox{1.\linewidth}{!}{
\begin{tabular}{lcccccc}
\toprule
\multirow{2}{*}{Method} &  \multicolumn{2}{c}{Distortion} & \multicolumn{4}{c}{Perceptual}  \\ \cmidrule(lr){2-3} \cmidrule(lr){4-7}
&  PSNR$\uparrow$ & SSIM$\uparrow$ & LPIPS$\downarrow$ & DISTS$\downarrow$ & FID$\downarrow$ & NIQE$\downarrow$  \\
\midrule
Baseline /w $\ell_1$ loss  & 23.21 & 0.826 & 0.114 & 0.109 & 51.99 & 4.67  \\
Baseline /w Entropy  & \textbf{24.05} & \textbf{0.848} & \textbf{0.081} & \textbf{0.071} & \textbf{37.20} & \textbf{4.40} \\

\bottomrule
\end{tabular}
}
\end{center}
\vskip -0.1in
\end{table}

\begin{table}[t]
\caption{Ablation experiment for the proposed entropy loss on LOL-v2-real~\cite{yang2020fidelity}. The Refusion~\cite{luo2023refusion} is used as the baseline.}
\label{table:entropy-lolv2}
\begin{center}
\resizebox{1.\linewidth}{!}{
\begin{tabular}{lcccccc}
\toprule
\multirow{2}{*}{Method} &  \multicolumn{2}{c}{Distortion} & \multicolumn{4}{c}{Perceptual}  \\ \cmidrule(lr){2-3} \cmidrule(lr){4-7}
&  PSNR$\uparrow$ & SSIM$\uparrow$ & LPIPS$\downarrow$ & DISTS$\downarrow$ & FID$\downarrow$ & NIQE$\downarrow$  \\
\midrule
Baseline /w $\ell_1$  & 18.54 & 0.812 & 0.151 & 0.136 & 51.99 & \textbf{3.98}  \\
Baseline /w Entropy  & \textbf{21.31} & \textbf{0.832} & \textbf{0.120} & \textbf{0.119} & \textbf{49.61} & 4.09 \\

\bottomrule
\end{tabular}
}
\end{center}
\vskip -0.1in
\end{table}

\subsection{NTIRE Low Light Enhancement Challenge}
\label{sec:ntire}

Our model was also evaluated in the NTIRE 2024 Low Light Enhancement Challenge~\cite{liu2024ntire_lle}, which provides a high-quality dataset containing 230 training scenes, along with 35 validation and 35 testing ones. We use the same model and settings as described in \cref{sec:implementation}. In this dataset, most images are 4K resolution that are hard to process with our diffusion model. Thus we downsample these large images with a factor of 0.5 for memory efficiency and resize them back to the original resolution after enhancement. The final results and rank are shown in~\Cref{table:ntire}, in which we choose the Top 10 teams and report our results in the last row. \textbf{Notably, our method outperforms all top teams in terms of LPIPS}.

\begin{figure*}[t]
  \centering
   \includegraphics[width=1.\linewidth]{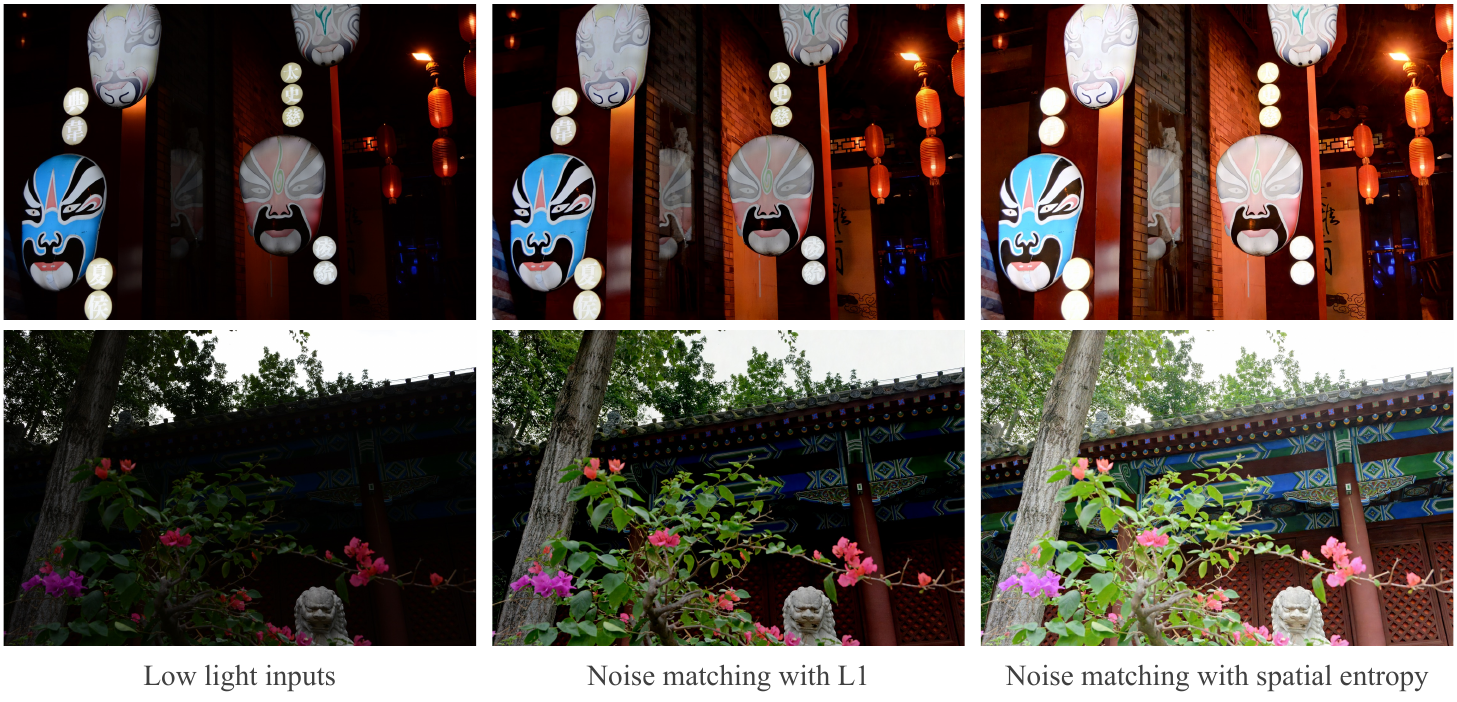}
   \vspace{-0.2in}
   \caption{Visual results of the proposed method on the NTIRE 2024 Low Light Enhancement Challenge dataset.}
   \label{fig:results_ntire}
   \vspace{-0.05in}
\end{figure*}

In~\Cref{fig:results_ntire}, we visualize two examples from the challenge validation dataset and illustrate the results of two diffusion models: learn noise matching with $\ell_1$ and noise matching with the proposed spatial entropy. Although there are no ground truth images, it is easy to observe that spatial entropy improves the visual quality of diffusion models.

\begin{table}[!t]
\footnotesize
\centering
\caption{Evaluation and Rankings in the NTIRE 2024 Low Light Enhancement Challenge.}
\setlength{\tabcolsep}{6pt}
\begin{tabular}{ccccc}
\toprule
\textbf{Team}                & \textbf{PSNR} & \textbf{SSIM} & \textbf{LPIPS} & \textbf{Final Rank} \\ 
\midrule
SYSU-FVL-T2 & 25.52 & 0.8637        & 0.1221   & 1 \\
Retinexformer                & 25.30         & 0.8525        & 0.1424                  & 2                   \\
DH-AISP                      & 24.97         & 0.8528        & 0.1235                     & 3                   \\
NWPU-DiffLight               & 24.78         & 0.8556        & 0.1673                & 4                   \\
GiantPandaCV                 & 24.83         & 0.8474        & 0.1353                 & 5                   \\
LVGroup\_HFUT                & 24.88         & 0.8395        & 0.1371                 & 6                  \\
Try1try8                     & 24.49         & 0.8483        & 0.1359             & 7                   \\
Pixel\_warrior               & 24.74         & 0.8416        & 0.1514              & 8                   \\
HuiT                         & 24.13         & 0.8484        & 0.1436              & 9                   \\
X-LIME                       & 24.28         & 0.8446        & 0.1298              & 10                  \\ \midrule
221B (\textbf{Ours})                        & 22.04         & 0.8141        & {\color[HTML]{FE0000} \textbf{0.1084}}        & 17   \\               
\bottomrule
\end{tabular}
\label{table:ntire}
\end{table}

\begin{figure*}[ht]
  \centering
   \includegraphics[width=1.\linewidth]{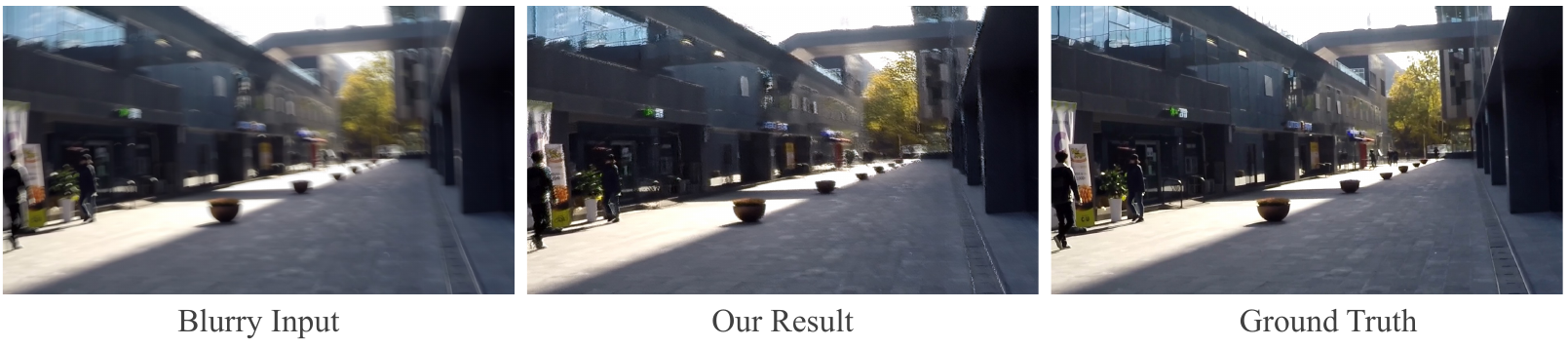}
   \vspace{-0.2in}
   \caption{A case of applying the proposed spatial entropy for image deblurring. Here we use a simple U-Net as the base network and directly train it with our differentiable spatial entropy loss.}
   \label{fig:deblur}
   \vspace{-0.05in}
\end{figure*}

\subsection{Limitation}
\label{sec:limitation}
Although the proposed spatial entropy can achieve good perceptual performance, the training is computationally costly since we need to count the pixel numbers for all bandwidths repeatedly. Therefore, we have to use a small patch size in training, which might affect the final performance. In addition, it is worth noting that the proposed entropy loss can be also used for other tasks and frameworks. However, since the loss is purely based on statistical feature matching, it reduces the weight of fidelity parts thus leading an inferior results on other metrics like PSNR and SSIM. That means directly applying it to some scenes would produce undesired artifacts. An example of applying it to image deblurring is shown in~\Cref{fig:deblur}. Although this problem can be alleviated using diffusion models in their iterative denoising process, it is still challenging to apply it to common image restoration frameworks. We regard this as our future work.

\section{Related Work}
\label{sec:related_work}

\paragraph{Low Light Image Enhancement} Low light enhancement aims to recover the correct illumination of images that are captured in the dark~\cite{land1977retinex,ooi2010quadrants}. To make the transformed images visually satisfactory, numerous works have made efforts to improve the contrast and details in restoration~\cite{fu2016weighted,singh2015enhancement,jobson1997multiscale,guo2016lime}. Traditional methods mainly adopt histogram equalization (HE)~\cite{bovik2010handbook,ooi2010quadrants,singh2015enhancement} and Retinex theory~\cite{land1977retinex,fu2016weighted}. Built upon them, deep learning approaches are developed and have achieved impressive progress. LLNet~\cite{lore2017llnet} is the first work applying deep neural networks to low light enhancement. Following this, RetinexNet~\cite{wei2018deep} proposes a real-world captured low light dataset with a Retinex theory based network. To improve the details, EnlightenGAN~\cite{jiang2021enlightengan} combines GAN loss with unpaired images for practical model training. And some other concurrent works mostly adopt similar architectures or traditional theories for data-driven image enhancement~\cite{guo2020zero,liu2021retinex,xu2022snr,wu2022uretinex,wang2019underexposed,zhang2021beyond,zhang2019kindling,Cai_2023_ICCV}. Most recently, DiffLL~\cite{jiang2023low} further utilizes a wavelet-based diffusion model to produce illuminated results with satisfactory perceptual fidelity.

\paragraph{Diffusion Models for Image Restoration}
Image restoration aims to reconstruct a high-quality image from its corrupted counterpart~\cite{andrews1977digital,luo2022deep,lian2022sliding,lian2023kernel}. It contains a wide range of practical applications such as image deblurring~\cite{shan2008high}, denoising~\cite{zhang2017beyond}, super-resolution~\cite{dong2016accelerating}, etc. Existing deep learning based methods directly train neural networks with $\ell_1$/$\ell_2$ loss based on collected data pairs, which is effective for image reconstruction but would produce over-smooth results~\cite{zhao2016loss,ledig2017photo}. Recently, the generative diffusion model has drawn increasing attention due to its stable training process and remarkable performance in producing realistic images and videos~\cite{ho2020denoising,saharia2022photorealistic,rombach2022high}. Inspired by it, recent researchers started converting various image restoration tasks into diffusion processes to obtain high-perceptual results~\cite{croitoru2023diffusion,luo2023controlling,kawar2022denoising,luo2023image,zhu2023denoising}. Notably, all these methods still use $\ell_1$/$\ell_2$ for noise matching to learn the diffusion process. In this paper, our method is the first that proposes to use a purely statistical matching approach for noise matching.
\section{Conclusion}
\label{sec:conclusion}

This paper presents a statistic-based matching approach (spatial entropy loss) for image restoration. Specifically, we introduce the kernel density estimation (KDE) to make the spatial entropy differentiable. Then the spatial entropy can be used for different learning-based frameworks for image reconstruction. By equipping it into the diffusion models (to substitute the $\ell_1$ or $\ell_2$), we obtain a novel statistical noise matching loss for realistic image restoration. We then apply this model to the low light enhancement task to illustrate its effectiveness. Our model achieves the best LPIPS performance in the NTIRE Low Light Enhancement challenge. All these results demonstrate that the spatial entropy loss is effective for the high perceptual diffusion learning process.

\noindent \textbf{Acknowledgements}
The computations were enabled by the \textit{Berzelius} resource provided by the Knut and Alice Wallenberg Foundation at the National Supercomputer Centre.

{
    \small
    \bibliographystyle{ieeenat_fullname}
    \bibliography{main}
}


\end{document}